\documentclass{article}
\pdfoutput=1
\usepackage[preprint]{neurips_2023}

\usepackage[utf8]{inputenc} 
\usepackage[T1]{fontenc}    
\usepackage{hyperref}       
\usepackage{url}            
\usepackage{booktabs}       
\usepackage{amsfonts}       
\usepackage{nicefrac}       
\usepackage{microtype}      
\usepackage{xcolor}         

\usepackage{amsmath,amsfonts,bm}

\def\eqref#1{equation~\ref{#1}}

\def\1{\bm{1}}

\def\vh{{\bm{h}}}

\DeclareMathAlphabet{\mathsfit}{\encodingdefault}{\sfdefault}{m}{sl}
\SetMathAlphabet{\mathsfit}{bold}{\encodingdefault}{\sfdefault}{bx}{n}

\def\gN{{\mathcal{N}}}

\newcommand{\E}{\mathbb{E}}

\usepackage{ellipsis}
\usepackage{wrapfig}
\usepackage{multicol}
\usepackage{multirow}
\usepackage{mwhmacros}
\usepackage{blindtext}
\usepackage{siunitx}
\usepackage{mathabx}
\usepackage{booktabs}
\usepackage{wasysym}
\usepackage{amsmath}
\usepackage{amssymb}
\usepackage{algorithm}
\usepackage{algorithmicx}
\usepackage{algpseudocode}
\usepackage{listings}
\usepackage{mathtools}
\usepackage{subfig}
\usepackage{geometry}
\usepackage{graphicx}
\usepackage[colorinlistoftodos]{todonotes}
\graphicspath{{./figs/}}
\usepackage[capitalise,nameinlink,noabbrev]{cleveref}
\title{Continuous Control Reinforcement Learning: \\Distributed Distributional DrQ Algorithms}

\newcommand{\drq}{{\bf DrQ-v2}}

\newcommand{\DDDrQ}{{\bf Distributed Distributional DrQ}}

\author{%
  ZHOU Zehao\thanks{Email: u3621365@connect.hku.hk} \\
  HKU\\
  Code: \url{https://github.com/ZHOU-henry/Distributed-Distributional-DrQ}\thanks{Original DrQ-v2 implementation based on ~\citep{yarats2021drqv2}}
}

\begin{document}

\maketitle
\newif\ifincludeappendix

\begin{abstract}
\DDDrQ{} is a model-free and off-policy RL algorithm for continuous control tasks based on the state and observation of the agent, which is an actor-critic method with the data-augmentation and the distributional perspective of critic value function. Aim to learn to control the agent and master some tasks in a high-dimensional continuous space. \drq{} use DDPG as the backbone and achieve out-performance in various continuous control tasks. Here \DDDrQ{} uses Distributed Distributional DDPG as the backbone, and this modification aims to achieve better performance in some hard continuous control tasks through the better expression ability of distributional value function and distributed actor policies.
\end{abstract}

\begin{figure}[h]
\centering
\def\mywidth{0.195}
\def\myhsep{-0.008}
\includegraphics[width=\mywidth\textwidth]{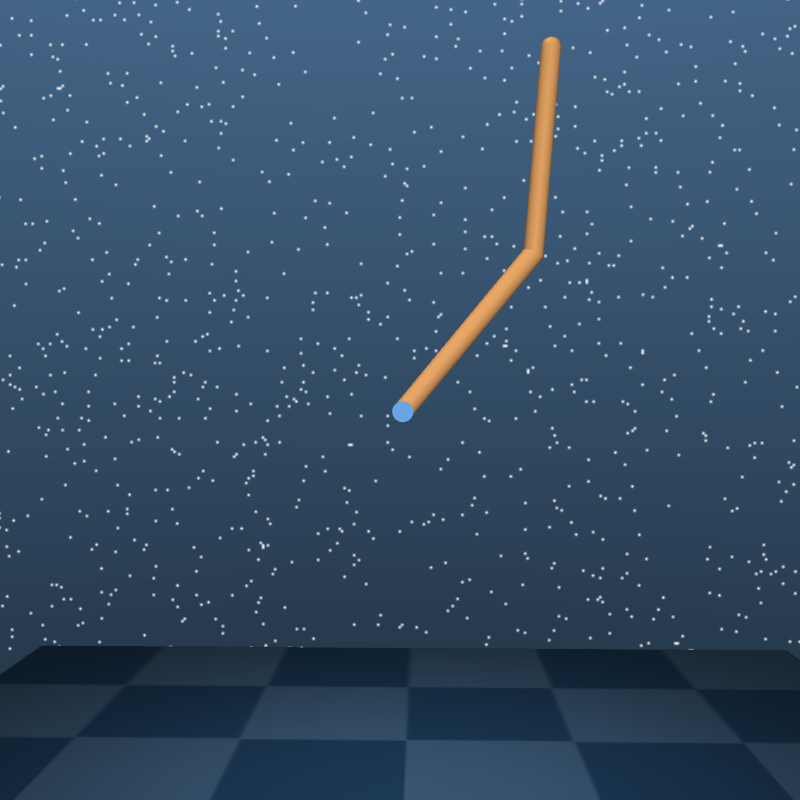}
\hspace{\myhsep\textwidth}
\includegraphics[width=\mywidth\textwidth]{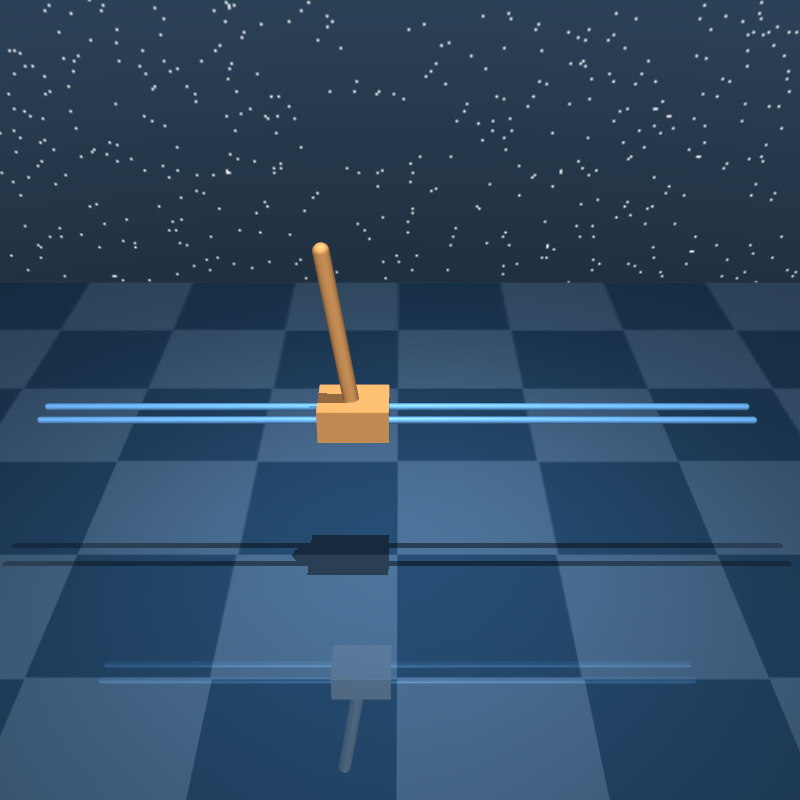}
\hspace{\myhsep\textwidth}
\includegraphics[width=\mywidth\textwidth]{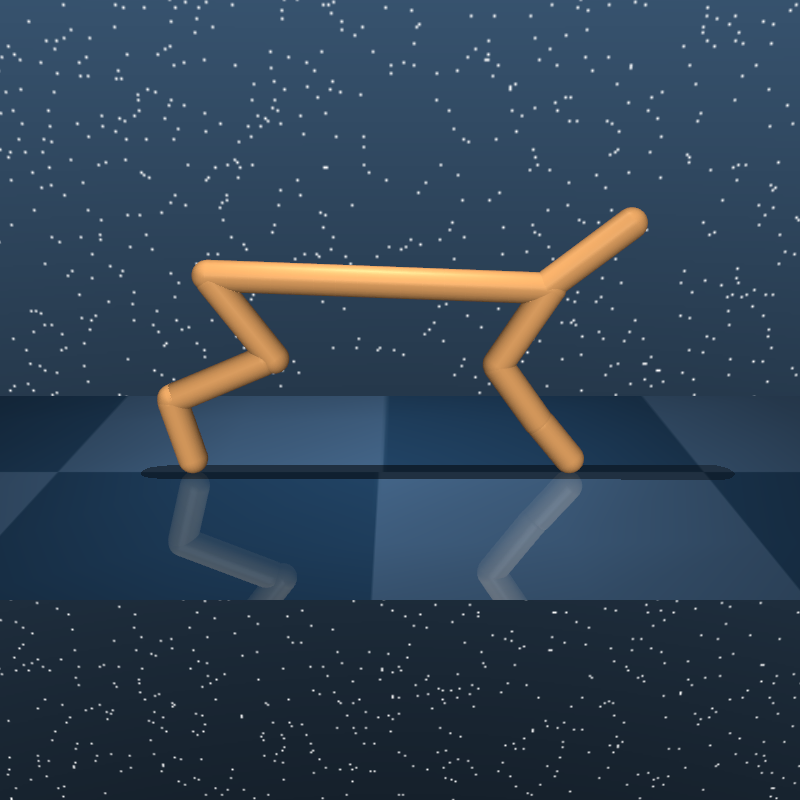}
\hspace{\myhsep\textwidth}
\includegraphics[width=\mywidth\textwidth]{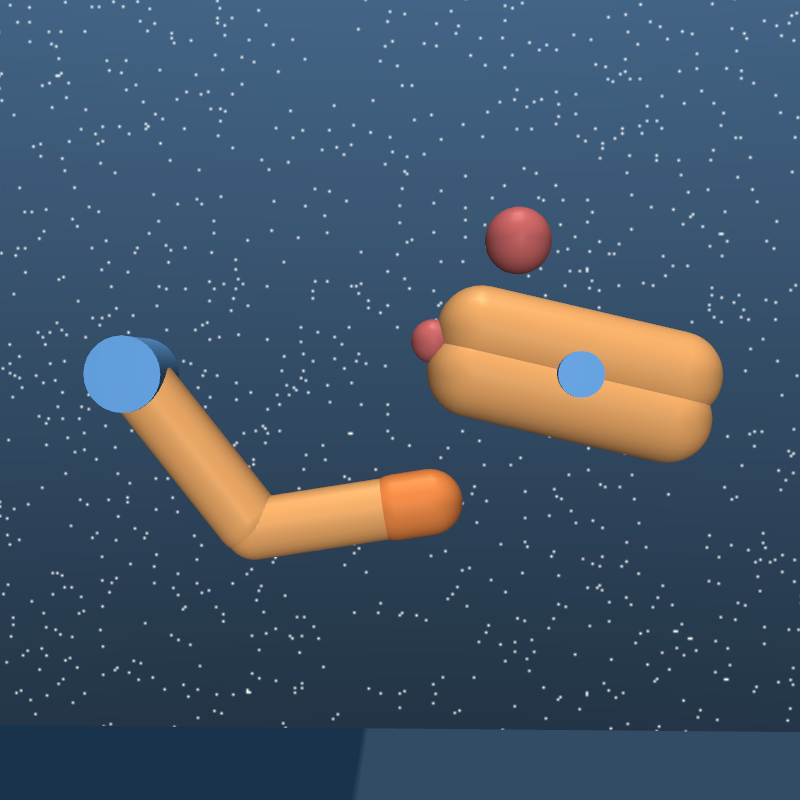}
\hspace{\myhsep\textwidth}
\includegraphics[width=\mywidth\textwidth]{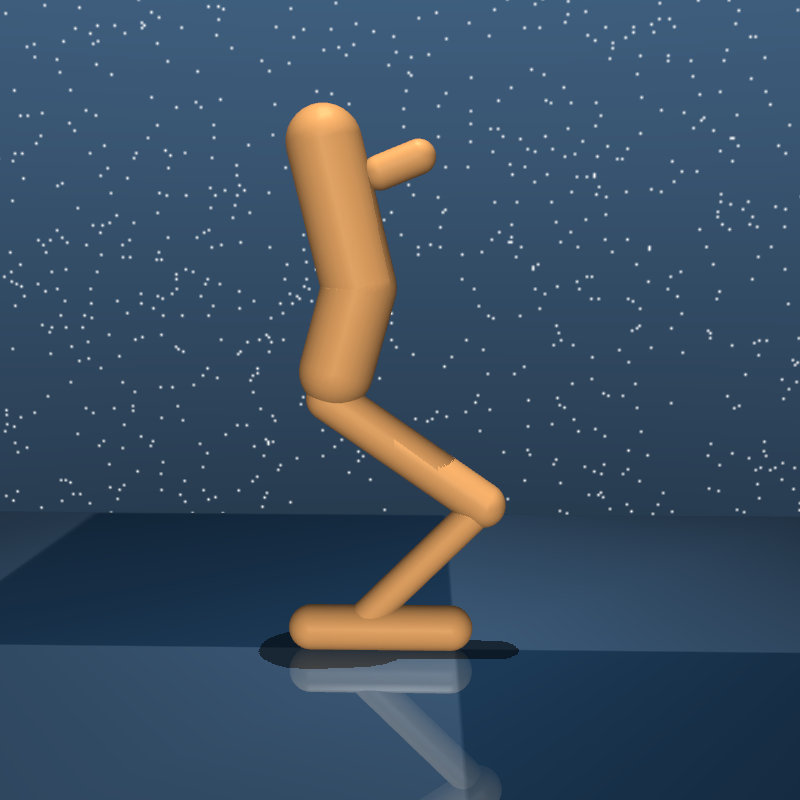}\\[0pt]
\includegraphics[width=\mywidth\textwidth]{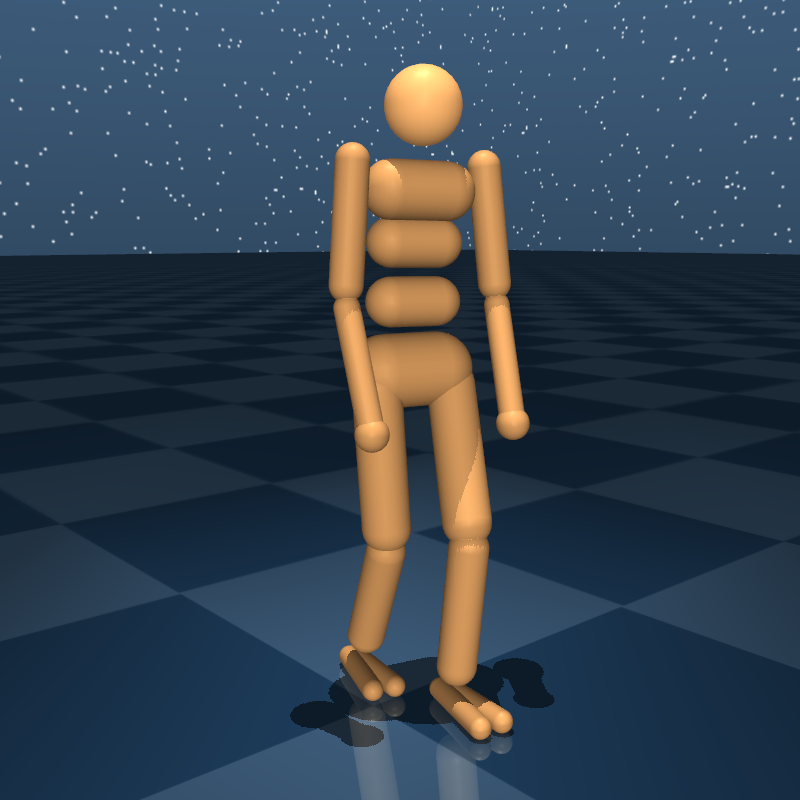}
\hspace{\myhsep\textwidth}
\includegraphics[width=\mywidth\textwidth]{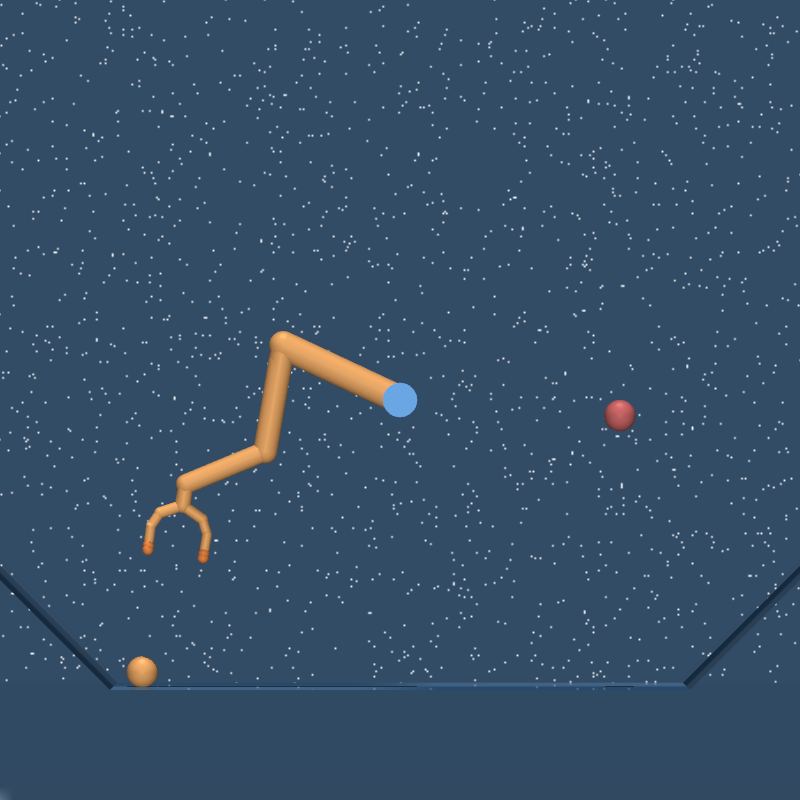}
\hspace{\myhsep\textwidth}
\includegraphics[width=\mywidth\textwidth]{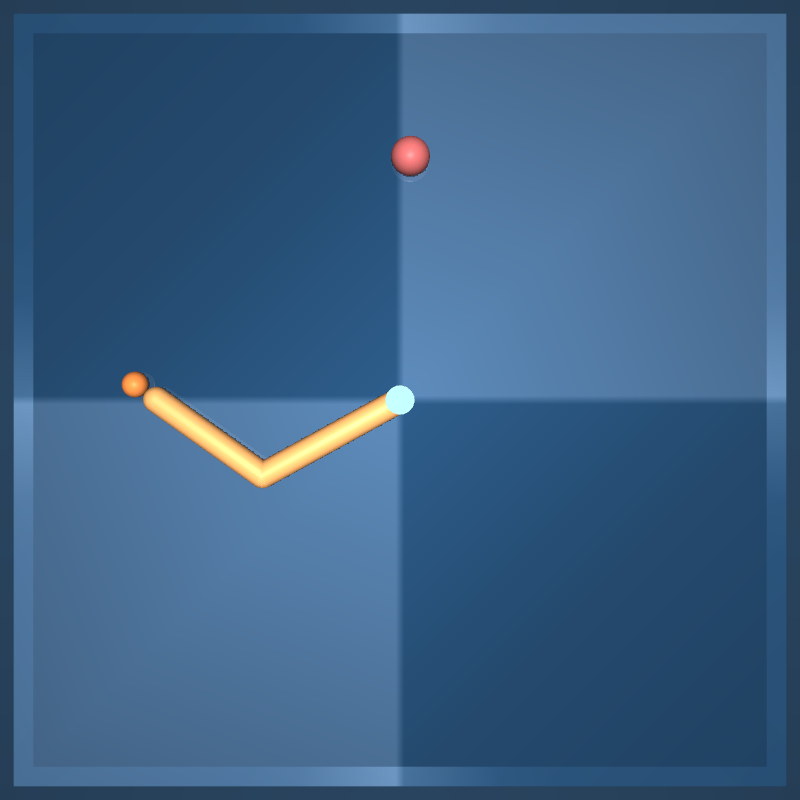}
\hspace{\myhsep\textwidth}
\includegraphics[width=\mywidth\textwidth]{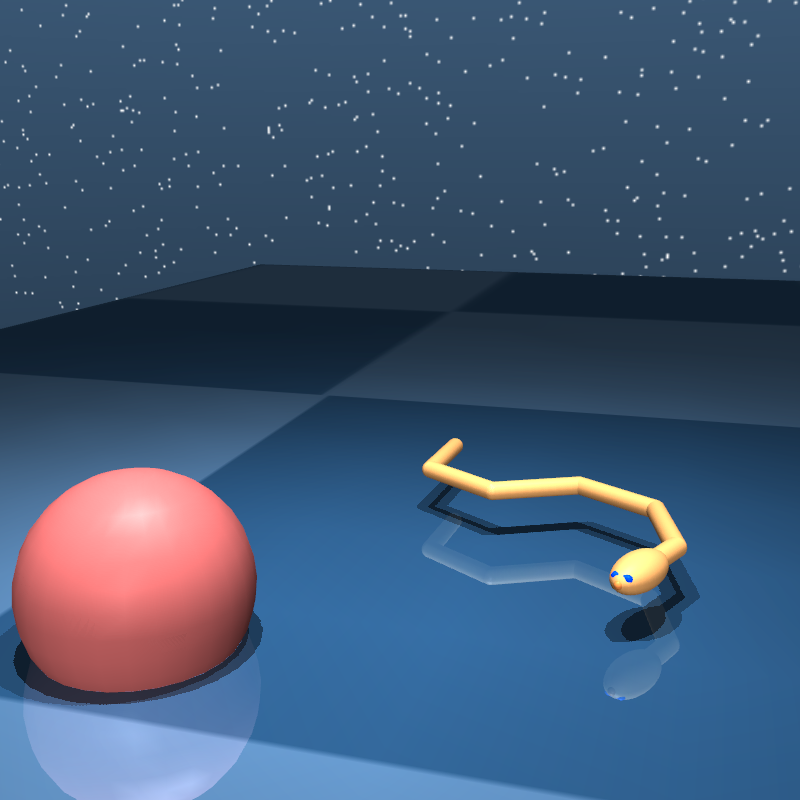}
\hspace{\myhsep\textwidth}
\includegraphics[width=\mywidth\textwidth]{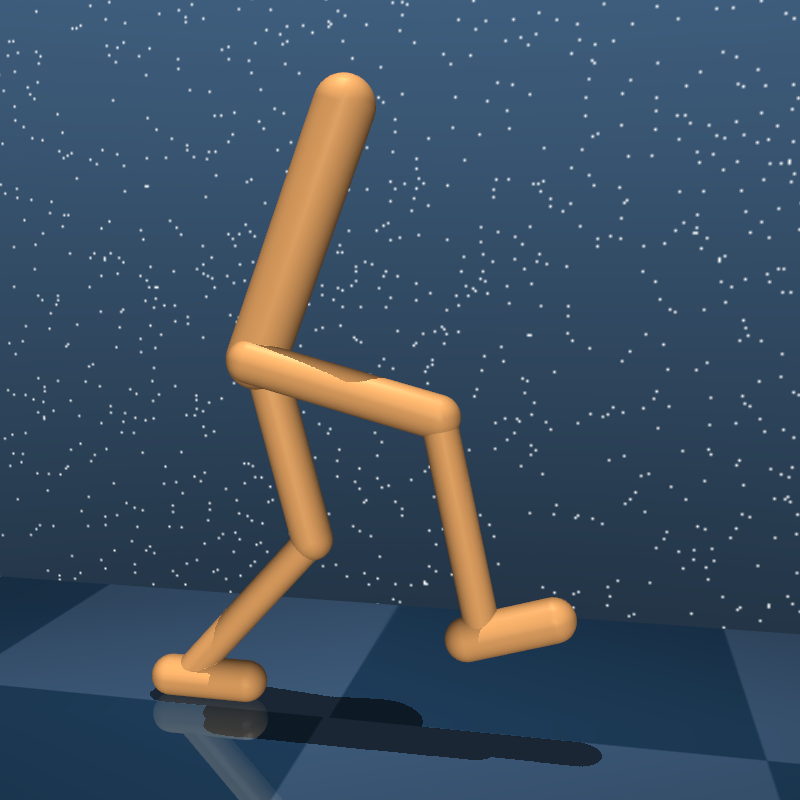}
\caption{DeepMind Control Suite: benchmark tasks. Domains include acrobot, cartpole, cheetah, finger, hopper, humanoid, manipulator, pendulum, reacher, swimmer6, walker. Picture from ~\citep{BarthMaron2018DistributedDD}}
\label{figure:1}
\end{figure}

\section{Introduction}
Reinforcement learning inspired by the neuroscience and animal learning theory, design rewards as guidelines for animal behavior. Optimal control and dynamic programming have been applied in real-world applications these decades ~\citep {Sutton1998}, and after combining with the deep learning method, deep Reinforcement learning (RL) started to master various challenging sequential decision-making and control problems. The high expressiveness of the artificial neural network makes deep Reinforcement learning become a possible model to deal with high-dimensional continuous environments and control tasks. As powerful non-linear function approximators, deep neural networks seem to have enough capacity to cover the complexity of high-dimensional data. Deep reinforcement learning showed human-level ability in the Atari video games with the algorithm of the Deep Q Network(DQN) ~\citep{mnih2013playing, Mnih2015HumanlevelCT}, and following reached a super-master level in the classical game of Go ~\citep{silver2016mastering, Silver2017MasteringTG}.

Before this, these competitive games were often considered to be very difficult for computers to defeat human experts because of their extremely large states and possibilities. Along with these discrete games, the machine learning community is also concerned about the problem of continuous control environments which are difficult to handle because of continuous infinite states and agent action space. However, continuous environments often occur in real-world problems such as robotics ~\citep{Gu2016DeepRL, Kalashnikov2018QTOptSD} and autonomous driving ~\citep{Bojarski2016EndTE}. In the continuous control tasks, it is a common choice to construct a long-term value function and then explicitly parameterize the action policy through function approximators and iterative modify the action policy according to the long-term approximation return of the value function. A sophisticated algorithm in the continuous control domain is deep deterministic policy gradient (DDPG) ~\citep{lillicrap2015ddpg}. Compared with on-policy deep reinforcement learning algorithms, such as Trust Region Policy Optimization(TRPO) ~\citep{Schulman2015TrustRP}, Proximal Policy Optimization(PPO) ~\citep{schulman2017proximal} or asynchronous advantage actor-critic (A3C) ~\citep{mnih2016asynchronous}, which need use online experiences as the bases of the approximation of value function at the same time with the real-time updating at each policy gradient step. The off-policy structure constructed through the Q-learning method ~\citep{Mnih2015HumanlevelCT} can use the samples from past stored experiences replay which is efficient when the complexity of tasks in the continuous control field increases. The off-policy improves the performance of the experience usage rate by repeatedly resampling the stored experience. However, when the RL algorithm combines non-linear Function approximation, bootstrapping, and off-policy training, the danger of instability and divergence arises which is called the deadly triad ~\citep{Sutton1998, Maei2009ConvergentTL}. Deep deterministic policy gradient in such cases becomes fragile and highly dependent on the selections of the hyper-parameters and random seeds ~\citep{Henderson2017DeepRL}, although after fine-tuning hyper-parameters, DDPG always produces good performance. DrQ-v2 as a model-free and off-policy deep RL algorithm uses DDPG as the backbone and combined Data-augmentation method shows out-performance in the continuous visual control tasks ~\citep{yarats2021drqv2}. DrQ-v2 shows sensitivity to the hyper-parameter and some random seeds in some continuous control tasks. To alleviate the vulnerability caused by DDPG, Distributed Distributional DrQ modified the action policy and critic value function of DrQ-v2 into the Distributional form of the objective function in Distributed Distributional DDPG(D4PG) ~\citep{BarthMaron2018DistributedDD, Bellemare2017ADP}, and the approximation distribution provides more information than the standard approximation of the single expectation value for the value function. The distributional perspective of the value function for MDP already has a history ~\citep{Jaquette1975MarkovDP}, such as in the Bayesian Q-Learning ~\citep{Dearden1998BayesianQ} to model parametric uncertainty through propagating the Q-values probability distributions, and Distributional value function shows play an important role in MDP and RL ~\citep{Bellemare2017ADP}.

\section{Related Work}

From history, the policy gradient method started from the likelihood ratio trick known as REINFORCE ~\citep{Williams2004SimpleSG}, the policy gradient method ~\citep{Sutton1999PolicyGM} inspire a promising concept Deterministic Policy Gradient (DPG) algorithm ~\citep{Silver2014DeterministicPG} which replaces the original stochastic policy with the deterministic policy, this modification is meaningful because it extended the deterministic policy gradient into model-free RL domain without integrate over whole action space and decreases the sample complexity. Later Deep Deterministic Policy Gradient (DDPG) ~\citep{Lillicrap2015ContinuousCW} combined DPG with deep neural network as the function approximators, and later Distributed Distributional DDPG (D4PG) ~\citep{BarthMaron2018DistributedDD} combined the distribution over returns ~\citep{Bellemare2017ADP} with DDPG algorithm. As a main RL method, Actor-critic algorithms use two separate functions: action policy and value functions to interact with environments and approximate future long-term returns respectively ~\citep{Konda1999ActorCriticA}. Distributed Distributional DrQ further combines several useful methods: visual-based representation learning ~\citep{doersch2015unsupervised, Wang_UnsupICCV2015} inspired the RL, convolution neural network in image ~\citep{Krizhevsky2012ImageNetCW} and auto-encoders ~\citep{finn2015deepspatialae} help to extract the feature from the continuous images in the visual RL domain; use action repeat, value function soft-update, and experiment replay ~\citep{Lin1992SelfimprovingRA, Mnih2015HumanlevelCT} to stabilize the deep neural network training; data-augmentation in DrQ-v2 ~\citep{yarats2021image} improve the data efficiency and improve the visual control performance, leverage soft actor-critic (SAC) ~\citep{Haarnoja2018SoftAO, Haarnoja2018SoftAA} and DDPG ~\citep{lillicrap2015ddpg} as the backbone objective function for the neural network optimization, and stack $3$ continuous RGB images of size $84 \times 84$ along the channel dimension which aim to capture the dynamic information of the agent in the specific environment states. Some other on-policy methods such as TRPO ~\cite{Schulman2015TrustRP} and PPO ~\citep{schulman2017proximal} use the synchronous policy gradient update method to search for better action policy. Some model-based reinforcement learning methods aim to construct the world model ~\citep{Ha2018WorldM} to capture the latent and abstract pattern of the discrete and continuous environment which the agent interact with, like PlaNet ~\citep{hafner2018planet}, dreamer ~\citep{Hafner2019DreamTC, Hafner2020MasteringAW, Hafner2023MasteringDD} and real-world implementation daydreamer ~\citep{Wu2022DayDreamerWM}. As another algorithm, Maximum a Posteriori Policy Optimization (MPO) based on the coordinate ascent on a relative entropy objective ~\citep{Abdolmaleki2018MaximumAP} also performs well in deep RL domain, and robustness to the hyper-parameter tuning in continuous control tasks, like standard Deepmind control suite ~\ref{figure:1} ~\citep{Tassa2018DeepMindCS, Tassa2020dmcontrolSA}.

\section{Preliminaries}

\subsection{Markov Decision Process \& Reinforcement Learning}

The current reinforcement learning community handles discrete and continuous sequential control tasks by constructing infinite-horizon or finite-horizon Markov Decision Process (MDP) ~\citep{Bellman1957AMD}. MDP describes an agent's interaction with an environment, the agent selects an action $a \in A$ based on the current state $x \in X$ at each time step, and the environment provides the corresponding reward $r \in R$ and the next states $x' \in X$. We can model these interactions as MDP $(X, A, R, P, \gamma)$. As conventional, $X$ and $A$ are states and actions space, $R$ and $P$ are the reward function and the transition function $P(\cdot | x, a)$, and $\gamma \in [0, 1]$ is the discount factor which aims to encourage the agent to gain reward as soon as possible. Furthermore, we define $\pi$ as a stationary action policy that maps each state $x \in X$ to an action probability distribution in the action space $A$. 

Without constraining the agent's behaviors with a certain explicit pattern, Reinforcement Learning aims to find the optimal policy $\pi$: $X \to A$ to maximize the expected value of the agent $Q$ ~\citep{Sutton1998}. Bellman's equation simply adds the expected value function $Q$ at the next action state and the expected reward that the agent gains during the interaction with the environment from current state $x, a$ to next state $(X', A')$, as $(x, a) \to (X', A')$. The value function under a certain policy $\pi$ can be written as $Q_\pi$ which describes the expected return with the policy $\pi$, and the value function $Q_\pi(x,a)$ is starting from selecting action $a \in A$ from state $x \in X$ and follow the policy $\pi$ in the following trajectory: 
\begin{align}
Q_\pi(x,a) = \E \left [ \sum_{t=0}^\infty \gamma^t R(x_t, a_t) \right ], x_t \sim P(\cdot | x_{t-1}, a_{t-1}), a_t \sim \pi(\cdot | x_t) \label{eqn:1}
\end{align}
According to Bellman’s equation ~\citep{Bellman1957AMD} and introduce the Bellman operator, the fundamental equation of MDP and RL: 

\begin{align}
(\calT_\pi Q)(x, a) = R(x, a) + \gamma \E\big[Q(x', \pi(x')) \big| x, a \big] \label{eqn:2}
\end{align}

The purpose of Reinforcement Learning is to find an optimal action policy $\pi^*$ for the agent in the environment to maximize the long-term reward (return):

\begin{align}
\pi^*: \E_{a \sim \pi^*} Q^*(x, a) = \max_a Q^*(x,a) \label{eqn:3}
\end{align}

These controls can be based on image input observations or proprioceptive state as input. In recent years, the combination of reinforcement learning and deep learning has greatly increased the handling ability of deep RL for some high-complexity environments and tasks. The deep neural network performs rich representation learning processing on the input images, encodes useful embedding for the control task from the high-dimensional data input, and achieves end-to-end control capabilities ~\citep{mnih2013playing, Mnih2015HumanlevelCT}. The goal of deep reinforcement learning is to find an optimal policy through learning in a given environment so that the agent can obtain the maximum expected long-term return from the interaction with the environment, deep RL leverages deep neural networks as action policy to increase the mapping capability in continuous spaces.

\subsection{Deep Deterministic Policy Gradient}

Deterministic Policy Gradient (DPG) algorithm ~\citep{Silver2014DeterministicPG} combined with deep neural network construct Deep Deterministic Policy Gradient (DDPG)\citep{lillicrap2015ddpg}, which is an off-policy, model-free actor-critic method with deterministic action policy. Based on Q-learning ~\citep{Watkins1992Qlearning}, the state-action value function is approximated as the expected long-term return followed by the action policy {/pi} starting from the state $x$ and action $a$. And through the sample experiences $(X, A, R, P, \gamma)$ from replay buffer dataset $D$, the value function keeps updating the expected return with the Bellman residual:

\begin{align}
L_\theta(D) = \E_{\substack{(x_t,a_t, r_t,x_{t+1}) \sim D }}(r_t + \gamma Q_{\theta '}(x_{t+1},\pi_\phi(x_{t+1})) - Q_\theta(x_t, a_t))^2 \label{eqn:4}
\end{align}
Action policy function $\pi_\phi(x_t)$ approximate to maximize the expected value function $Q_\theta(x_t, a)$: 
\begin{align}
L_\phi(D) = - \E_{x_t \sim D} Q_\theta(x_t, \pi_\phi(x_t)) \label{eqn:5}
\end{align}

\begin{algorithm}[t]
\caption{\DDDrQ: Distributional perspective}
\label{alg:DDDrQ}
\begin{algorithmic}
    \State \textbf{Inputs:} \\
        Auto-encoder $f_{encoder}$ and action policy $\pi_\phi$.\\
        Training steps $T$, Exploration scheduled standard deviation $\sigma(t)$.\\
        \State \textbf{Whole training procedure:}
        \For {$t=1...T$}
        \State Compute Exploration scheduled standard deviation and update: $\sigma(t)$
        \State Sample exploration noise from Normal distribution: $\epsilon \sim \gN(0, \sigma_t^2)$
        \State Add exploration noise to the deterministic action: $a_t \leftarrow \pi_\phi(f_{encoder}(x_{t})) + \epsilon$
        \State Obtain the new state from environment: $x_{t+1} \sim P(\cdot | x_t, a_t)$ 
        \State Collect and store the experience tuple to the replay buffer: $D \leftarrow D \cup (x_t, a_t, r_t, x_{t+1})$ 
        
        \State \textbf{Update Critic-function} ~\ref{alg:critic}
        \State \textbf{Update Actor-policy} ~\ref{alg:actor}
        
        \EndFor    
\end{algorithmic}
\end{algorithm}

\subsection{D4PG: Distributed version of DDPG}

Compared with the standard RL objective function, the Distributed Distributional DDPG algorithm (D4PG) ~\citep{BarthMaron2018DistributedDD} aims to learn an action policy $\pi$ from a distributional perspective of the value function. Distributional value function $Q_\pi(x, a)= \E Z_\pi(x, a)$ contains more information about the performance of the current action policy, this will provide a more accurate approximation for the following update and search for the action policy, the distributional Bellman operator can be defined as:

\begin{align}
    (\calT_\pi Z)(x, a) = r(x, a) + \gamma \E\big[ Z(x', \pi(x')) \big| x, a \big] \label{eqn:6}
\end{align}

\section{Distributed Distributional DrQ}
In the previous section, the Distributed Distributional DDPG algorithm (D4PG) ~\citep{BarthMaron2018DistributedDD} is shown as a practical off-policy method for action policy learning. In this section, we describe Distributed Distributional DrQ, an off-policy model-free actor-critic RL method which is based on the Distributed Distributional DDPG as the backbone and upon some data pre-process methods from DrQ-v2 ~\citep{yarats2021image} such as frame-stack and image-augmentation.

\subsection{Data preprocess}

Continuous control tasks are based on the visual image, proprioceptive, or both, the algorithm first passes the pixel observation data into the convolution neural network to auto-encoder into an abstract and low-dimensional latent space ~\citep{finn2015deepspatialae}. For increasing the data efficiency, during the update step, both the critic and actor functions use the image-augmentation data ~\citep{yarats2021image}. Image augmentation applies a random shift to the image pixels and selects a random $84 \times 84$ crop after padding 4 pixels, The augmentation function also replaces the pixel value with around nearest pixel values by the bilinear interpolation method. Image augmentation increases the data efficiency and accelerates the action policy convergence speed ~\citep{yarats2021drqv2}. During the update process, both the critic value function and action policy function use $h = f_{encoder}(\mathrm{augmentation}(x))$ as the observation input, $x$ and $h$ are the original image data and latent encoder vector respectively.

\subsection{Distributed Distributional Deep Deterministic Policy Gradient}

\begin{algorithm}[t]
\caption{\DDDrQ: Update Critic-function}
\label{alg:critic}
\begin{algorithmic}
    \State \textbf{Inputs:} \\
    Batch size $B$, Learning rate $\alpha$, soft update rate $\tau$, clip value $c$, action policy $\pi_\phi$.\\
    Auto-encoder $f_{encoder}$, random image-augmentation $\mathrm{aug}$ and critic value function parameters $\theta$.\\
    \State Sample a mini-batch of $B$ transitions: $\{(x_t, a_t, r_{t:t+n-1}, x_{t+n})\} \sim \mathcal{D}$
    \State Apply data augmentation and encode: $h_t, h_{t+n} \leftarrow f_{encoder} (\mathrm{aug}(x_t)), f_{encoder} (\mathrm{aug}(x_{t+n}))$ 
    \State Sample action: $a_{t+n} \leftarrow \pi_\phi(\vh_{t+n}) + \epsilon$ and $\epsilon \sim \mathrm{clip}(\gN(0, \sigma^2), -c, c)$
    \State Compute critic losses: $L_{\theta_1, {encoder}}$ and  $L_{\theta_2, {encoder}}$ using ~\eqref{eqn:8}
    \State Update critic weights: $\theta_k \leftarrow \theta_k -\alpha \nabla_{\theta_k} L_{\theta_k, {encoder}} , \quad \forall k \in \{1,2\}$
    \State Update critic target weights: $\bar{\theta}_k \leftarrow \tau \theta_k + (1-\tau)\bar{\theta}_k , \quad \forall k \in \{1,2\}$
    \State Update encoder weights: ${encoder} \leftarrow {encoder} -\alpha \nabla_{encoder} (L_{\theta_1, {encoder}} + L_{\theta_2, {encoder}})$
\end{algorithmic}
\end{algorithm}

\begin{figure}[h]
\begin{center}
\includegraphics[width=0.60\textwidth]{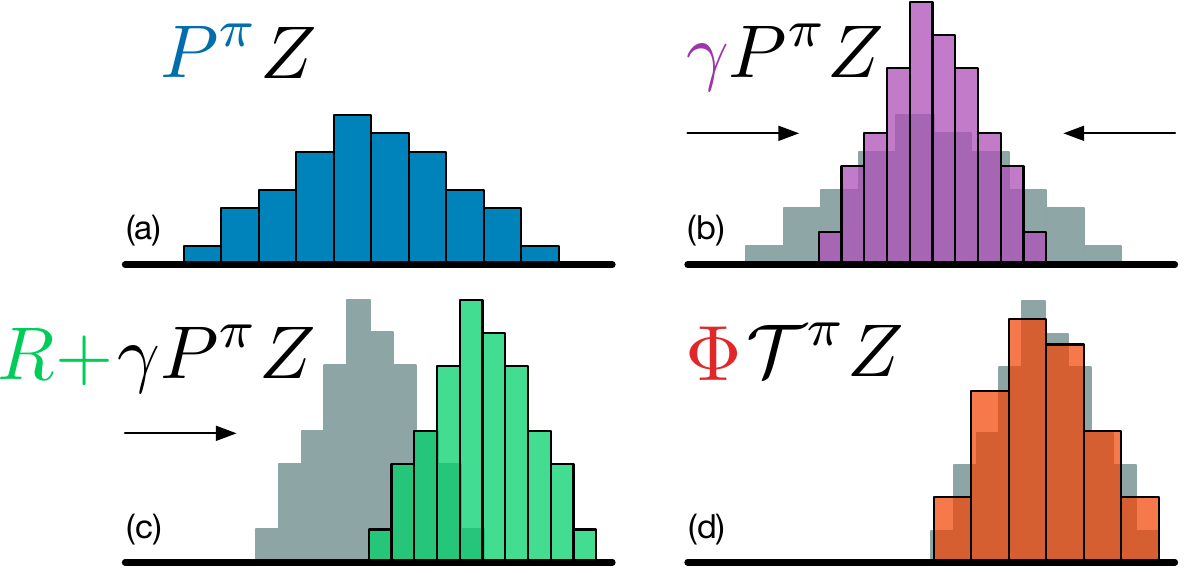}
\end{center}
\caption{A distributional perspective of value function: (a) one-step distribution under policy $\pi$, (b) add discount to the distribution, (c) add reward to the distribution (d) project the distribution. Picture from ~\citep{Bellemare2017ADP} \label{fig:distribution}}
\end{figure}

Distributed Distributional DrQ follows many efficient data processing methods from DrQ-v2 and changes the backbone actor-critic RL algorithm from DDPG ~\citep{lillicrap2015ddpg} to Distributed Distributional DDPG (D4PG) ~\citep{BarthMaron2018DistributedDD, Bellemare2017ADP}. The motivation for this modification is to reduce the sensitivity and fragility of the DDPG algorithm to hyper-parameters, aim to make the algorithm more robust to different hyper-parameters and random seeds, and the cost is the computational speed of the whole learning process. D3rQ applies double Q-learning to mitigate the overestimation bias for the critic value function ~\citep{Fujimoto2018AddressingFA}, which normally uses two separate value functions $Q_{\theta_1}$ and $Q_{\theta_2}$. And approximate the target value function and TD error with n-step return ~\citep{Sutton1998}. Multi-step returns make the whole learning process performance more stable and increase the reward propagation efficiency. For this, we sample a mini-batch of agent transitions experiences $\tau =(x_t, a_t, r_{t:t+n-1}, x_{t+n})$ from the replay buffer $D$, like defined above ~\eqref{eqn:1} and ~\eqref{eqn:2} and then calculate the target critic function through the N-step distributional Bellman operator ($Q_{\theta_k}(x, a) = \E\big[Z_{\theta_k}(x, a)\big]$) which can be defined as:

\begin{align}
(\calT_\pi ^n Z) (x_t, a_t) = \sum_{i=0}^{n-1} \gamma^i r_{t+i} + \gamma^n \min_{k=1,2} \E\big[ Z_{\bar \theta_k}(x_{t+n}, \pi(x_{t+n})) \big| x_t, a_t \big] \label{eqn:7} \end{align}

\begin{algorithm}[t]
\caption{\DDDrQ: Update Actor-policy}
\label{alg:actor}
\begin{algorithmic}
    \State \textbf{Inputs:} \\
    Batch size $B$, Learning rate $\alpha$, soft update rate $\tau$, clip value $c$, action policy $\pi_\phi$.\\
    Auto-encoder $f_{encoder}$, random image-augmentation $\mathrm{aug}$ and actor policy parameters $\phi$.\\
    \State Sample a mini-batch of $B$ observations: $\{(x_t)\} \sim \mathcal{D}$
    \State Apply data augmentation and encode: $h_t \leftarrow f_{encoder}(\mathrm{aug}(x_t))$
    \State Sample exploration noise from Normal distribution: $\epsilon \sim \mathrm{clip}(\gN(0, \sigma^2), -c, c)$ 
    \State  Add noise to the action: $a_{t} \leftarrow \pi_\phi(h_{t}) + \epsilon$
    \State Compute actor loss: $L_{\phi}$ using ~\eqref{eqn:9}
    \State Update actor's weights: $\phi \leftarrow \phi -\alpha \nabla_\phi L_{\phi}$
\end{algorithmic}
\end{algorithm}

Above is the N-step distributional variant of the standard one-step Bellman operator ~\eqref{eqn:6}, and the expectation of the distributional value function contains N-step transition information which increases the training stability and reward propagation speed. Similar to the MSE loss function in ~\eqref{eqn:4}, with some distance metrics $d$ between two distributions, we can define the distributional loss function:

\begin{align}
L_{\theta}(D) = \E_{(x_t,a_t, r_t,x_{t+1}) \sim D } \Big[ d((\calT_{\pi_{\theta'}} ^n Z) (x, a), Z_{\theta}(x, a)) \Big] \label{eqn:8}
\end{align}

Combine the above distributional value function with the standard DDPG action policy $\pi_\phi(x_t)$ gradient update ~\eqref{eqn:5}, the action policy gradient is constructed by updating forward the direction maximizing the action-value distribution:

\begin{align}
L_\phi(D) = - \E_{x_t \sim D}\Big[\nabla_\phi \pi_\phi(x)\,\E \nabla_{a} Z_{\theta}(x, a) \big|_{a=\pi_\phi(x)}\Big] \label{eqn:9}
\end{align}

\subsection{Categorical distribution}

\begin{figure}[t]
    \centering
    \includegraphics[width=0.7\linewidth]{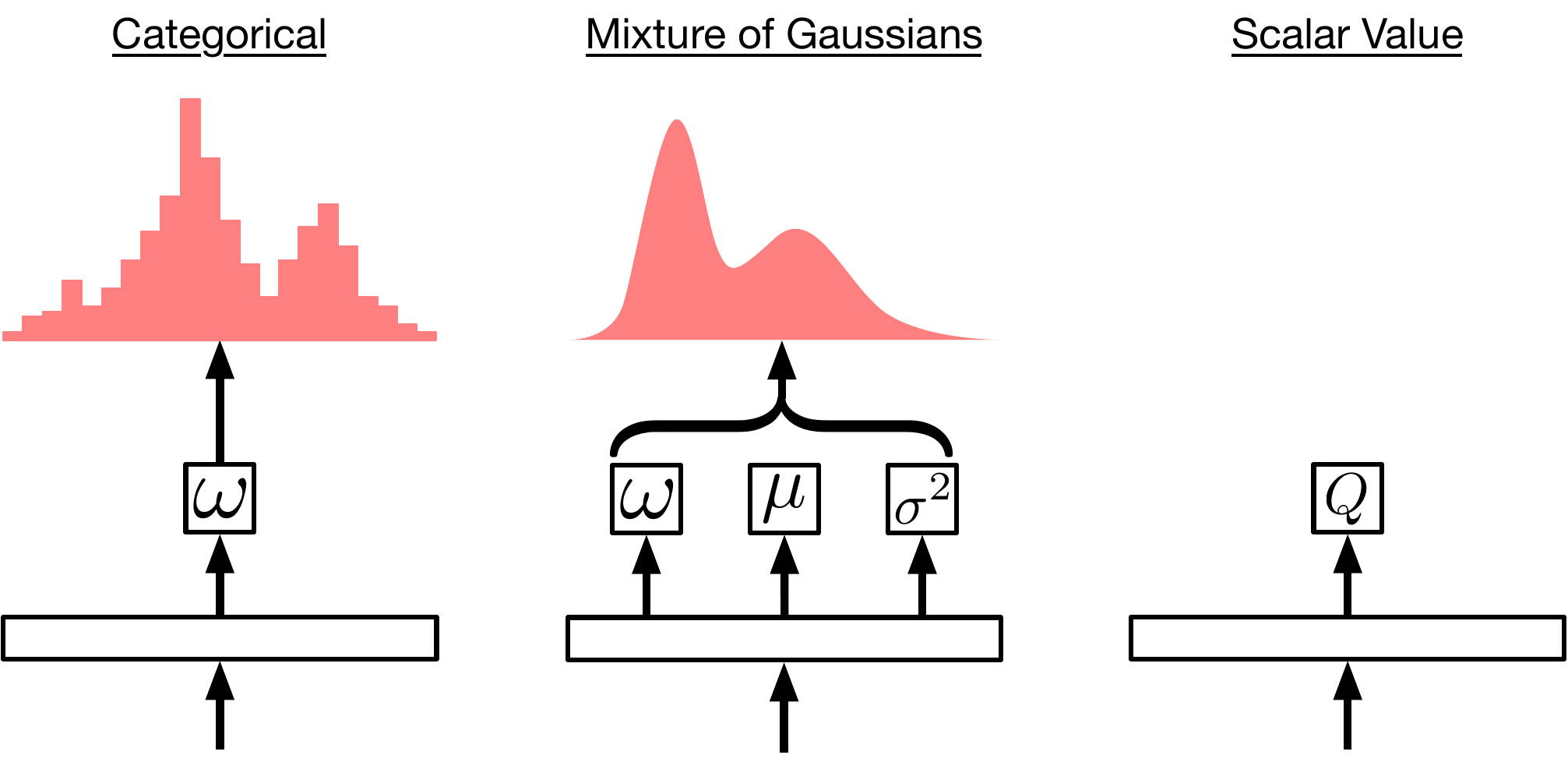}
    \caption{above are the output parameters of different value function types for the last layer of the critic neural networks, including the categorical distribution, mixture of Gaussians distribution, and standard scalar value output. Picture from ~\citep{BarthMaron2018DistributedDD}}
    \label{fig: distribution}
\end{figure}

Following the original paper (D4PG) ~\citep{BarthMaron2018DistributedDD} using the categorical distribution and the mixture of Gaussians as the distribution type for the value function ~\ref{fig: distribution}, we use categorical distribution with the bounds ($V_{\min}, V_{\max}$) and the number of atoms $n$ as the hyper-parameters, and $\Delta = \frac{V_{\max} - V_{\min}}{n-1}$ corresponds to the distance between two atoms, and then for each atom position $y_i = V_{\min} + i\Delta$. For each $z_i$, $\omega_i$ is the output parameter which represents the value of the categorical distribution at that atom. We can then define the action-value distribution as
\begin{align}
    Z= z_i \quad \text{w.p.} \quad p_i \propto \exp\{\omega_i\}.
\end{align}
Followed by a softmax activation function used for normalization:
\begin{align}
p_i = \sigma (\omega_i) = \frac{\exp\{\omega_i\}}{\sum_j \exp\{\omega_j\}}
\end{align}

then for the loss function for the distributional value function with the cross-entropy loss ~\citep{Bellemare2017ADP}:
\begin{align}
d(\Phi\calT_{\pi_\phi} Z_\mathrm{target}, Z) = \sum_{i=0}^{n-1} p_i' \frac{\exp\{\omega_i\}}{\sum_j \exp\{\omega_j\}}.
\end{align}

\section{Conclusion}

The distributional value function as the critic value function provides more fruitful information for the update of actor policy, this makes the whole policy gradient method process more stable, robust, and accurate compared with the algorithm using standard DDPG as the backbone. Furthermore, we need to notice that D4PG costs more computational resources compared with standard DDPG at each step and this reduces the fps during the training period. This is because D4PG constructs more neurons at the last two layers and categorical distribution projection also needs more computation. Therefore, as a kind of trade-off, D3rQ commonly needs more computation time than DrQ-v2 for processing the same amount of frames and parameter updates.

\bibliographystyle{plainnat}

\includeappendixtrue 
\ifincludeappendix
\newpage
\appendix

\section{Hyper-parameters}

\begin{table}[hb!]
\caption{Default setting of Hyper-parameters}
\centering
\begin{tabular}{lc}
\hline
Parameter        & Setting \\
\hline
Agent update frequency & $2$ \\
Optimizer & Adam \\
Learning rate & $10^{-4}$ \\
Mini-batch size & $256$ \\
Discount $\gamma$ & $0.99$ \\
Action repeat & $2$ \\
$n$-step returns & $3$ \\
Exploration steps & $2000$ \\
Replay buffer capacity & $10^6$ \\
Exploration stddev. clip & $0.2$ \\
Critic Q-function soft-update rate $\tau$ & $0.01$ \\
Features dim. & $50$ \\
Hidden dim. & $1024$ \\
Exploration stddev. schedule & $\mathrm{linear}(1.0, 0.05, 200000 \to 3000000)$ \\
\hline
\end{tabular}
\end{table}

\section{Mujoco Control suite details}
\begin{table}[hb!]
    \caption{Domains and tasks in the Mujoco Control Suite ~\citep{Tassa2018DeepMindCS, Tassa2020dmcontrolSA}.}
    \centering
    \begin{tabular}{llccc} \toprule
    Domain                    & Task            & $|\mathcal{A}|$     & $|\mathcal{S}|$     & $|\mathcal{X}|$     \\ \midrule
    \multirow{2}{*}{acrobot}  & swingup         & \multirow{2}{*}{1}  & \multirow{2}{*}{4}  & \multirow{2}{*}{6}  \\
                              & swingup\_sparse &                     &                     &                     \\ \midrule
    \multirow{2}{*}{cartpole} & swingup         & \multirow{2}{*}{1}  & \multirow{2}{*}{4}  & \multirow{2}{*}{5}  \\
                              & swingup\_sparse &                     &                     &                     \\ \midrule
    cheetah                   & walk            & 6                   & 18                  & 17                  \\ \midrule
    \multirow{2}{*}{finger}   & turn\_easy      & \multirow{2}{*}{2}  & \multirow{2}{*}{6}  & \multirow{2}{*}{12} \\
                              & turn\_hard      &                     &                     &                     \\ \midrule
    \multirow{2}{*}{fish}     & upright         & \multirow{2}{*}{5}  & \multirow{2}{*}{27} & \multirow{2}{*}{24} \\
                              & swim            &                     &                     &                     \\ \midrule
    hopper                    & stand           & 4                   & 14                  & 15                  \\ \midrule
    \multirow{3}{*}{humanoid} & stand           & \multirow{3}{*}{21} & \multirow{3}{*}{55} & \multirow{3}{*}{67} \\
                              & walk            &                     &                     &                     \\
                              & run             &                     &                     &                     \\ \midrule
    manipulator               & bring\_ball     & 2                   & 22                  & 37                  \\ \midrule
    \multirow{2}{*}{swimmer}  & swimmer6        & 5                   & 16                  & 25                  \\
                              & swimmer15       & 14                  & 34                  & 61                  \\ \bottomrule
    \end{tabular}
    \label{table:tasks}
\end{table}


\end{document}